\documentclass[article, 10 pt, conference]{ieeeconf}  
\IEEEoverridecommandlockouts                              

\UseRawInputEncoding 
\let\labelindent\relax
\usepackage{enumitem}
\setlist{topsep=1pt, partopsep=0pt, parsep=0pt, itemsep=0pt}
\usepackage[
top    = 0.75in,
bottom = 0.75in,
left   = 0.75in,
right  = 0.75in]{geometry}
\usepackage{cite}
\usepackage{amsmath}
\usepackage[skip=10pt plus1pt, indent=35pt]{parskip}
\usepackage{placeins}
\usepackage{textcomp}
\usepackage[dvipsnames]{xcolor}
\definecolor{royalblue}{RGB}{65, 105, 225}
\definecolor{maroon}{RGB}{180, 0, 0}
\definecolor{DarkGreen}{RGB}{0, 100, 0}
\usepackage{booktabs}
\usepackage{multirow}
\usepackage{siunitx}
\usepackage{algorithm}
\usepackage{algpseudocode}
\usepackage{graphicx}
\usepackage[symbol]{footmisc}
\usepackage{adjustbox}
\usepackage{float}
\usepackage{svg}
\usepackage{array}
\usepackage[T1]{fontenc}
\usepackage{subcaption}
\usepackage{xspace}
\usepackage{scrextend}
\usepackage{caption}
\usepackage{tabularx}
\usepackage{listings}
\usepackage{wrapfig}
\usepackage{calligra}
\usepackage{comment}
\usepackage{soul}
\usepackage{balance}
\usepackage{listings}
\usepackage{xcolor}
\usepackage{caption}
\newcolumntype{A}{ >{\centering\arraybackslash} m{4cm} }
\newcolumntype{B}{ >{\centering\arraybackslash} m{1cm} }
\newcolumntype{C}[1]{>{\centering\let\newline\\\arraybackslash\hspace{0pt}}m{#1}}

\makeatletter
\newcommand\footnoteref[1]{\protected@xdef\@thefnmark{\ref{#1}}\@footnotemark}
\makeatother

\usepackage{amsmath,lipsum} 
\usepackage{amssymb}  
\usepackage{amsfonts}
\usepackage{textgreek}
\usepackage{authblk}
\usepackage{etoolbox}
\usepackage{algpseudocode}

\algrenewcommand\algorithmicindent{0.5em}

\makeatletter
\let\NAT@parse\undefined

\let\oldthebibliography\thebibliography
\let\endoldthebibliography\endthebibliography
\renewenvironment{thebibliography}[1]{
  \oldthebibliography{#1}
  \setlength{\itemsep}{-1.75ex plus-.1ex minus-.1ex} 
}{
  \endoldthebibliography
}

\makeatother

\usepackage[colorlinks=true, citecolor=cyan]{hyperref}  
\def\BibTeX{{\rm B\kern-.05em{\sc i\kern-.025em b}\kern-.08em
    T\kern-.1667em\lower.7ex\hbox{E}\kern-.125emX}}

\title{\LARGE \bf
AdaptBot: Combining LLM with Knowledge Graphs and Human Input for Generic-to-Specific Task Decomposition and Knowledge Refinement}

\author{ Shivam Singh$^{1*}$, Karthik Swaminathan$^{1*}$, Nabanita Dash$^1$, Ramandeep Singh$^1$ \\  Snehasis Banerjee$^2$,  Mohan Sridharan$^3$,  Madhava Krishna$^1$
\thanks{*Denotes equal contribution}

\thanks{$^{1}$ Robotics Research Center, IIIT Hyderabad, India}
\thanks{$^{2}$ TCS Research, Tata Consultancy Services, India}
\thanks{$^{3}$ School of Informatics, University of Edinburgh, UK}
}

\makeatletter
\renewcommand{\@seccntformat}[1]{%
  \protect\csname the#1\endcsname\protect\quad%
}
\makeatother

\renewcommand{\thesection}{\arabic{section}}
\renewcommand{\thesubsection}{\thesection.\arabic{subsection}}
\renewcommand{\thesubsubsection}{\thesubsection.\arabic{subsubsection}}

\begin{document}

\maketitle
\thispagestyle{empty}
\pagestyle{empty}

\begin{abstract}
An embodied agent assisting humans is often asked to complete new tasks, and 
there may not be sufficient time or labeled examples to train the agent to perform these new tasks. Large Language Models (LLMs) trained on considerable knowledge across many domains can be used to predict a sequence of abstract actions for completing such tasks, although the agent may not be able to execute this sequence due to task-, agent-, or domain-specific constraints. Our framework addresses these challenges by leveraging the generic predictions provided by LLM and the prior domain knowledge encoded in a Knowledge Graph (KG), enabling an agent to quickly adapt to new tasks. The robot also solicits and uses human input as needed to refine its existing knowledge. Based on experimental evaluation in the context of cooking and cleaning tasks in simulation domains, we demonstrate that the interplay between LLM, KG, and human input leads to substantial performance gains compared with just using the LLM. \\
Project website\footnote[4]{Project supported in part by TCS Research India}: \href{https://sssshivvvv.github.io/adaptbot/}{https://sssshivvvv.github.io/adaptbot/}

\end{abstract}
\vspace{-1em}
\begin{keywords}
Large Language Models, Knowledge Graph, Human-in-the-loop Learning
\end{keywords}

\section{Introduction}
\label{sec:introduction}
Embodied agents are being used in assistive roles in many applications, aided in part by the availability of realistic simulators~\cite{coppeliaSim,Puig_2018_CVPR,kolve2022ai2thorinteractive3denvironment}. Although such agents possess some prior knowledge of domain objects and their attributes, they are often asked to perform new tasks and operate in new scenarios. For example, an agent preparing dishes in the kitchen based on prior knowledge of some recipes and ingredients, may be asked to prepare a new dish or clean the pantry.


\vspace{-0.75em}
Large Language Models (LLMs) trained on a large corpus of data have demonstrated the ability to decompose a range of tasks into a sequence of high-level (abstract) actions (i.e., sub-tasks) that implement the task~\cite{khot2023decomposedpromptingmodularapproach,reppert2023iterateddecompositionimprovingscience,liu2024deltadecomposedefficientlongterm}. For example, an LLM can provide a sequence of sub-tasks for completing the previously unseen task of \textit{preparing hot chocolate}. However, this sequence may involve incorrect steps, or reference objects and actions that the agent does not have access to in the kitchen under consideration.

\begin{figure}[tb]
\centering
\captionsetup{font=scriptsize}
\setlength{\belowcaptionskip}{-10pt}
\includegraphics[width=0.49\textwidth]{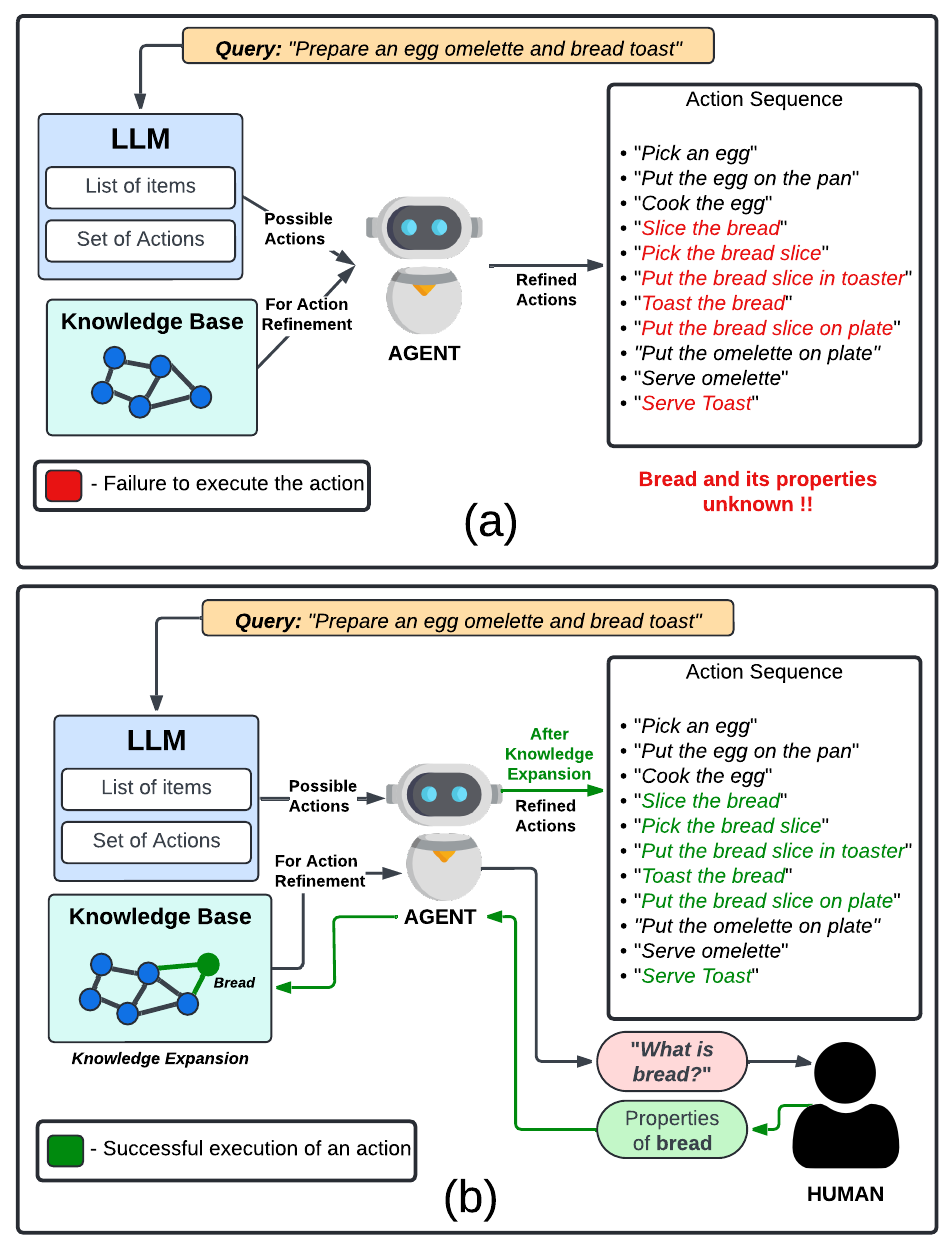}
\vspace{-0.5em}
\caption{For any given task, an LLM provides a generic sequence of abstract actions that is refined using the domain-specific knowledge in a KG. If the sequence refers to objects, attributes, or actions that cannot be resolved using the KG, or leads to unexpected outcomes, human input helps refine or expand the KG.}
\vspace{-6pt}
\label{fig:teaser}
\end{figure}

\vspace{-0.75em}
The challenges mentioned above are partially offset by the fact that an assistive agent usually has some prior domain-specific knowledge in the form of objects, object attributes, and action capabilities. State-of-the-art methods build large datasets of such information for a given application domain~\cite{sakib2022approximate}, or attempt to embed this knowledge by repeatedly tuning deep networks~\cite{sakib2024cooking}. However, such knowledge is not readily available for many practical domains, and modern data-driven methods make it difficult to reliably and transparently revise the encoded knowledge over time. In a departure from such methods, the framework described in this paper seeks to leverage the complementary strengths of LLMs, Knowledge Graphs (KGs), and human feedback---see Figure~\ref{fig:teaser}. Our framework enables the assistive agent to:

\begin{figure*}[h]
  \centering
  \captionsetup{font=scriptsize}
  \includegraphics[width=\textwidth]{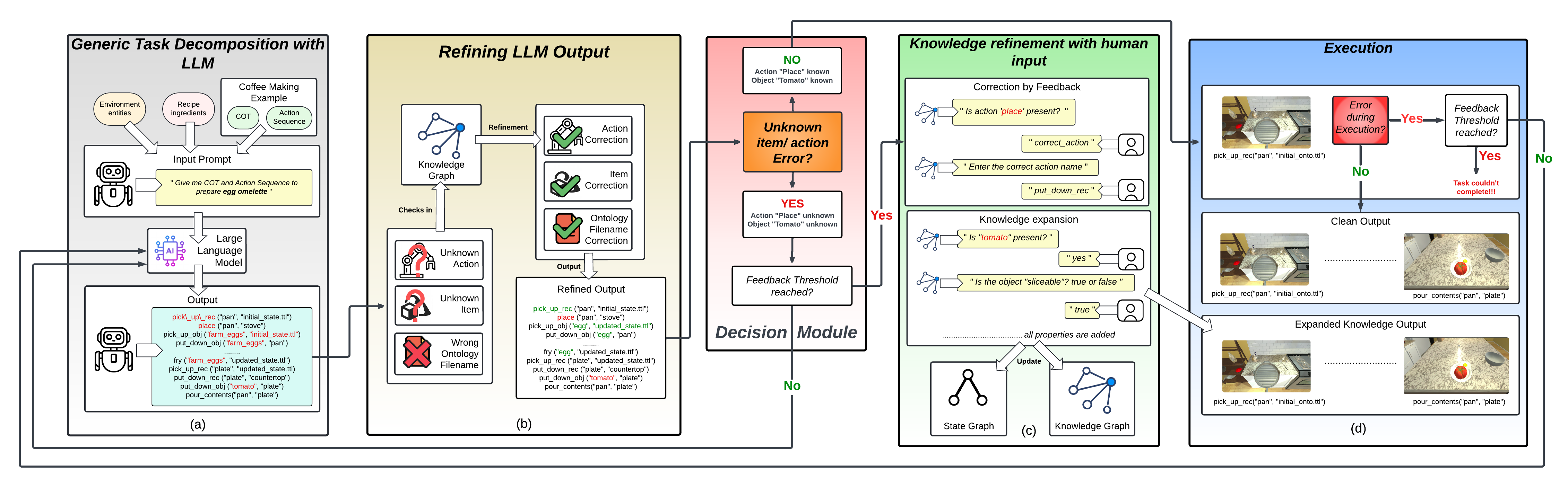}
  \vspace{-2em}
  \caption{Framework overview for cooking tasks: (a) Input Chain-of-Thought (COT) prompt contains target dish, available ingredients, and an example of input and output action sequence (for task of making coffee), to obtain an output action sequence; (b) Any mismatch (e.g., in object classes, actions) between LLM output and KG are identified and action sequence is revised if possible; (c) Agent attempts to resolve any remaining errors or unexpected outcomes by re-prompting LLM, with errors that persist being addressed by soliciting human input and updating KG; (iv) Revised/corrected action sequence is executed.}
  \label{fig:pipeline}
  \vspace{-1em}
\end{figure*}

\begin{enumerate}
    \vspace{-1em}
    \item Query an LLM to obtain a generic sequence of actions (sub-tasks) to be executed to accomplish any given task.
    \item Encode any prior domain-specific knowledge of object types and attributes in a KG, using it to revise the LLM's output action sequence.
   
    \item Use discrepancies between LLM output, KG, and observations of action outcomes to support human-in-the-loop (HITL) refinement of the knowledge in the KG.
    \vspace{-1.5em}
\end{enumerate}
We illustrate and evaluate these capabilities in two different classes of tasks: cooking and cleaning, demonstrating: (a) substantial improvement in performance compared with baselines that use just the LLM or even a combination of LLM and KG for completing an assigned task; and (b) the ability to adapt to new classes of tasks through incremental knowledge refinement instead of elaborate tuning (e.g., of LLMs) or encoding comprehensive knowledge.

\vspace{-0.75em}
The remainder of the paper is organized as follows. We begin with a discussion of related work (Section~\ref{sec:related-work}), followed by a description of our proposed framework (Section~\ref{sec:framework}). We then discuss the experimental set up and results (Section~\ref{sec:exp-setup-results}), followed by the conclusions (Section~\ref{sec:conclusions}).






\section{Related Work}
\label{sec:related-work}
We motivate our framework by discussing related work in the use of LLMs, KGs, and HITL task decomposition.

\textbf{LLMs and KGs for task decomposition}: LLMs such as GPT-4~\cite{openai2024gpt4technicalreport}, Gemma2~\cite{gemmateam2024gemma2improvingopen}, and LLaMA3~\cite{dubey2024llama3herdmodels} have experimentally demonstrated the ability to decompose abstract tasks into sub-tasks~\cite{khot2023decomposedpromptingmodularapproach, reppert2023iterateddecompositionimprovingscience, wen2024learning,li2023semantically,dery2021auxiliarytaskupdatedecomposition,liu2024deltadecomposedefficientlongterm}. Frameworks such as TaskBench~\cite{shen2023taskbench} have compared fully automated processes with those with human interventions, particularly for unfamiliar or "open-set" tasks~\cite{wang2024tdag, cui2021semantic,liu2024deltadecomposedefficientlongterm}. Additionally, methods such as ADaPT~\cite{prasad2023adapt} have supported iterative adjustment of task complexity continuously based on real-time feedback. In parallel, KGs have have been used to model prior knowledge of objects and their attributes for sequential task planning, e.g., for sequential task prediction with graph CNN~\cite{article1}, action planning for robots in Industry 4.0 environments~\cite{10.1145/3297280.3297568}, and for generalizing to new (related) environments ~\cite{9561782}. Our framework builds on these ideas by combining the (generic) prediction capabilities of LLMs with the domain-specific knowledge encoded in a KG~\cite{aburasheed2024knowledgegraphscontextsources,pan2024unifying} for task adaptation in new environments~\cite{kuang2024openfmnavopensetzeroshotobject}.


\vspace{-0.75em}
\textbf{Related task planning examples}: The Functional Object-Oriented Network (FOON)~\cite{paulius2016functional} encodes substantial knowledge about cooking (e.g., ingredients and outcomes of actions) in the form of task trees and using them for task planning for cooking related dishes~\cite{ding2022robottaskplanningsituation,bhat2024groundingllmsrobottask,jiang2019task,sakib2022approximate}. In more recent work, a fine-tuned GPT has been used to transform generic recipe instructions into task trees, which are merged and revised by comparing information stored in FOONs to obtain the task tree used for execution~\cite{sakib2024cooking}. 
These methods use examples from the Recipe1M+ dataset~\cite{marin2021recipe1m+} for tuning and evaluation. Instead of tuning an LLM across classes of tasks or training a knowledge base extensively for a particular class of tasks, our framework supports incremental revision, faster adaptation, and reliability. Our framework provides the assistive agent limited (prior) knowledge of any specific domain as a KG, enabling it to incrementally refine the KG with new objects and actions as they are encountered, and to correct errors by soliciting and using human feedback when it is necessary and available. 

\vspace{-0.75em}
\textbf{Human-in-the-loop task decomposition}: Human feedback has been used to enhance hierarchical task allocation and robot task planning in complex environments~\cite{marzari2021towards,zhen2023robottaskplanningbased}. Frameworks like TaskBench~\cite{shen2023taskbench} and Reflexion~\cite{shinn2024reflexion} leverage human feedback to iteratively decompose tasks, making LLMs more effective in handling abstract tasks. Hierarchical task structuring is crucial for handling complex, multi-step task decomposition, especially in abstract problem domains~\cite{holler2020hddl}. Instead of iteratively tuning LLMs (e.g., through prompts), which does not necessarily lead to correct results, we combine the generic prediction capabilities of LLM, real-time domain-specific KG updates~\cite{DING2019105,article}, and human-in-the-loop feedback~\cite{kasaei2024vitalvisualteleoperationenhance,liu2023robotlearningjobhumanintheloop,9044335,emami2024human,wu2022survey},  allowing the system to operate based on the available knowledge to perform new classes of tasks while incrementally refining the knowledge.

\section{Problem Formulation and Framework}
\vspace{-0.5em}
\label{sec:framework}
Figure~\ref{fig:pipeline} is an outline of our framework. In the motivating example, an agent assisting in cooking tasks in a kitchen has access to relevant objects and ingredients for many dishes but it does not have the recipes. When asked to prepare any particular dish, $\tau_i$, the agent queries an LLM to obtain a sequence of abstract actions (sub-tasks), i.e., $\langle a_1, \ldots, a_{m_i}\rangle$. For example, the sequence for \textit{make an omelette} includes \textit{picking up the egg} and \textit{breaking the egg over a skillet}. This sequence of abstract actions is checked against a KG with some domain-specific information in the form of existing objects and attributes that include the actions that can be performed on some objects. The agent tries to resolve any discrepancy between the LLM output and KG, e.g., KG states there is no skillet or that an egg can only be cracked, by finding replacements, e.g., \textit{crack the egg over a pan}. If the discrepancy is not resolved, or if executing the action sequence does not provide the desired outcome, the agent identifies relevant actions and solicits human input to refine the KG, e.g., add knowledge of objects or their attributes, and provides an action sequence to complete the task. The agent is assumed to be able to execute these actions. We describe out framework's components below.

\subsection{Generic Task Decomposition with LLM}
\label{sec:framework-llm}
In our framework, we use an LLM to decompose any given task into a sequence of sub-tasks because LLMs have demonstrated the ability to provide such a sequence of abstract actions for many different tasks. Specifically, in the motivating example, the LLM is prompted with information about some domain objects, an example cooking task (make coffee), and the corresponding action sequence (recipe) to be executed---see Figure~\ref{fig:pipeline}a. We experimentally evaluate the use of different LLMs, as described in Section~\ref{sec:expres-setup}. 

\vspace{-0.75em}
Since the sequence of sub-tasks predicted by the LLM is based on many information sources, it may not be possible to execute one or more of these actions. For example, in the context of cooking tasks, the suggested ingredient may not be available or the action may involve an incorrect choice of tool (e.g., using a fork to cut vegetables). These situations can be addressed in part by using prior domain-specific information, which is encoded as described below.

\subsection{Representing Domain-specific Knowledge with KG}
\label{sec:framework-kg}
Our framework uses a Knowledge Graph (KG) to encode any prior information available to the agent. In the context of cooking tasks, this includes knowledge of some classes of ingredients (e.g., herbs, fruits, vegetables), receptacles (e.g., plates, bowls, countertop), and tools (e.g. knives, spoons), which can be arranged hierarchically. It also encodes the existence of some specific instances of these object classes and their properties such as likely location(s) and the actions they can be involved in (e.g., cutting, scooping, grinding). We use the \textit{Resource Description Framework} (RDF) format to encode this information in two graph structures in Turtle format (.ttl file)---see Figure~\ref{fig:kg-nodes}:
\begin{enumerate}
\vspace{-0.75em}
\item \textbf{State graph:} models current state as $\mathbf{G_s} = (\mathbf{I_s}, \mathbf{E_s})$, where nodes $\mathbf{I_s}$ are instances of object classes such as ingredients and receptacles; and $\mathbf{E_s} \subseteq \mathbf{I_s} \times \mathbf{P_s} \times \mathbf{V_s}$ are edges such that $(i_j, p, v_k) \in \mathbf{E_s}$ is a triple denoting an attribute of $i_j \in \mathbf{I_s}$ in terms of value $v_k \in \mathbf{V_s}$ of predicate $p \in \mathbf{P_s}$. For example, (apple1, obj\_location, fridge) and (apple1, is\_sliced, true) express \textit{apple1's} location and that it is sliced.
\item \textbf{Attribute graph:} encodes the known properties and action capabilities of some object classes as $\mathbf{G_k} = (\mathbf{I_k}, \mathbf{E_k})$, where nodes $\mathbf{I_k}$ represent the classes and edges $\mathbf{E_k} \subseteq \mathbf{I_k} \times \mathbf{P_k} \times \mathbf{V_k}$ represent class properties, e.g., (apple, sliceable, true) implies  apples can be sliced.
\vspace{-0.75em}
\end{enumerate}
The available actions include moving, picking up, and putting down objects; using tools; cleaning, toggling, slicing, stirring, and mopping\footnote{Supplementary material includes list of all the actions.}. Such a KG can be learned automatically based on information extracted from datasets or sensor streams. The feasibility of any action/sub-task in the sequence predicted by LLM is then checked using $\mathbf{G_k}$ and $\mathbf{G_s}$ by generating suitable SPARQL queries. If the predicted sequence of actions passes the KG-based check, it is executed, changing $\mathbf{G_s}$ suitably.
\begin{figure}[tb]
\captionsetup{font=scriptsize}
\centering
    \begin{minipage}{0.3\textwidth}
        \begin{lstlisting}[basicstyle=\ttfamily\scriptsize]
ex:onion rdf:type ex:object ;
    ex:obj_name 'onion' ;
    ex:IsSliceable true ;
    ex:Fryable true ;
    ex:NeedsToBeCleaned true .
        \end{lstlisting}
    \end{minipage}
    \hfill

    \begin{minipage}{0.3\textwidth}
        \begin{lstlisting}[basicstyle=\ttfamily\scriptsize]
ex:onion rdf:type ex:object ;
    ex:obj_name 'onion' ;
    ex:obj_location ex:fridge .
    ex:sliced false ;
    ex:IsFried false ;
    ex:IsCleaned false .
        \end{lstlisting}
    \end{minipage}
    \setlength{\abovecaptionskip}{-2pt}
    \setlength{\belowcaptionskip}{-12pt}
    \caption{Example of a node \textit{onion} in $\mathbf{G_k}$ (\textit{top}) and $\mathbf{G_s}$ (\textit{bottom}).}
    \label{fig:kg-nodes}
    \vspace{-0.5em}
\end{figure}


\subsection{Refining LLM output}
\label{sec:framework-refine}
If a mismatch is detected between the LLM output and the KG, the agent attempts to use the KG to \textit{revise} the action sequence---see Figure~\ref{fig:pipeline}(b). 
Specifically, the agent attempts to replace the text corresponding to the identified mismatch, which can refer to actions, object instances, or object attributes, with other text from the KG.
While performing such text replacement, it is important to consider syntactic similarity, which measures similarity in the structure (e.g., of words or sentences), and semantic similarity, which considers similarity in meaning.
In our framework, the agent can compute the similarity of the identified words (or their embedding) with words from a similar category (or their embedding) in the KG. The use of word embeddings requires additional contextual information and makes it difficult to understand the revision of the LLM output. We thus chose to use the direct matching of words while considering hypernyms (broader terms) or hyponyms (more specific terms) for simplicity, ease of use, and transparency. If the agent is able to replace all identified mismatches, it executes the actions. 

\subsection{Knowledge refinement with human input}
\label{sec:framework-hitl}
Since the KG is not comprehensive, the agent may not be able to resolve all identified mismatches, e.g., reference to unknown object or action. Also, there may be unexpected action outcomes when the agent executes the action sequence. These situations are handled through re-prompting and human feedback---see Figure~\ref{fig:pipeline}(c).
Specifically, the agent responds to an unresolved mismatch or erroneous outcome by re-prompting the LLM with additional information (of mismatch or error). If the mismatch or error persists, human input is solicited and used. 

\vspace{-0.75em}
\noindent
\textbf{Existence check:} if an action or object in the LLM output does not exist in the KG, there are three possibilities: (1) The agent is mistaking an existing item (action) for another item (action); (2) the entity does not exist in the domain; or (3) the entity exists but is not in the KG. In the first case, human informs the agent about the correct object (or action); in the second case, human denies existence of entity; and in the third case, human confirms the entity's existence and agent interactively obtains entity's attributes\footnote{Supplementary material includes details of questions asked.}. As the agent expands its knowledge, the need for human input progressively decreases.

\vspace{-0.75em}
\noindent
\textbf{Learn attributes:} If human confirms existence of an instance of a new entity, the agent interactively obtains additional details. For example, when informed about an instance of a new object class \textit{onion}, agent incrementally requests information about the object type (e.g., \textit{edible\_object}) and other relevant attributes (e.g., boilable, fryable, location of instance). This knowledge revision can be viewed as correcting (expanding) the knowledge in the KG by revising class attributes in $\mathbf{G_k}$ and instance-specific details in $\mathbf{G_s}$.
$$f_{KE}(I_{new}, P_{new}, S_{current}) \Rightarrow {\mathbf{G_{k}^{'}}, \mathbf{G_{s}^{'}}}$$
where $I_{new}$ is the new entity; $P_{new}$ = ${(p_1, v_1), ..., (p_n, v_n)}$ refers to attributes ($p_i$) of entity and their values ($v_i$); $S_{current}$ = ${(s_1, v_1), ..., (s_n, s_n)}$ refers to states $s_i$ and their values $v_i$; and $\mathbf{G^{'}_{k}}$ and $\mathbf{G^{'}_{s}}$ are the updated components of the KG.  For example, new edge is added in $\mathbf{G_s}$ to encode an onion's position and new edge is added in $\mathbf{G_k}$ to encode that an onion can be fried. Note that this update to existing knowledge is fully transparent by design.


    

\begin{algorithm}[tb]
\caption{LLM + KG + Human Input}
\label{alg:algorithm}
\begin{algorithmic}[1]
\State \textbf{Procedure} LLM\_KG\_Human($\mathbf{G_{s}}$, $\mathbf{G_{k}}$, \textit{ip\_prompt})
    \State $F$ $\leftarrow 0$                     \algorithmiccomment{$F$ is feedback counter}
    \State $T$ $\leftarrow$ call\_LLM(\textit{ip\_prompt}) \algorithmiccomment{$T$ is action sequence}
    \State $T_{refined}$, $\varepsilon_{unkn}$ $\leftarrow$ refine\_sequence($T$, $\mathbf{G_{k}}$, $\mathbf{G_{s}}$)
    \If {NOT $\varepsilon_{unkn}$}                \algorithmiccomment{$\varepsilon_{unkn}$ is unknown\_item error}
        \State $O$, $\varepsilon_{exec}$ $\leftarrow$ execute($T_{refined}$) \algorithmiccomment{$O$ is execution output}
        \null\hfill\algorithmiccomment{$\varepsilon_{exec}$ is execution error}
    \EndIf

    \While {($\varepsilon_{exec}$ OR $\varepsilon_{unkn}$) AND $F$ $<$ $F_{max}$}
        
        \While {$\varepsilon_{unkn}$ AND $F$ $<$ $F_{max}$}
            \State $T^{'}$ $\leftarrow$ call\_LLM(fb\_prompt) \algorithmiccomment{$T^{'}$is updated sequence}
            \State $T^{'}_{refined}$, $\varepsilon_{unkn}$ $\leftarrow$ refine\_sequence($T^{'}$, $\mathbf{G_{k}}$, $\mathbf{G_{s}}$)
            \State $F$ $\leftarrow$ $F$ + 1
        \EndWhile

        \If {$\varepsilon_{unkn}$ AND $F$ == $F_{max}$}
            \State $T^{'}_{refined}$ $\leftarrow$ ask\_human($\varepsilon_{unkn}$) \algorithmiccomment{$\mathbf{G_{k}}, \mathbf{G_{s}} \Rightarrow \mathbf{G_{k}^{'}}, \mathbf{G_{s}^{'}}$}
            \State $O$, $\varepsilon_{exec}$ $\leftarrow$ execute($T^{'}_{refined}$)
            \State \textbf{break}
        \EndIf

        \If {$\varepsilon_{exec}$ AND $F$ $<$ $F_{max}$}
            \State $T^{'}$ $\leftarrow$ call\_LLM(fb\_prompt)
            \State $T^{'}_{refined}$, $\varepsilon_{unkn}$ $\leftarrow$ refine\_sequence($T^{'}$, $\mathbf{G_{k}}$, $\mathbf{G_{s}}$)
            \State $F$ $\leftarrow$ $F$ + 1
        \EndIf

        \If {NOT $\varepsilon_{unkn}$}
            \State $O$, $\varepsilon_{exec}$ $\leftarrow$ execute($T^{'}_{refined}$)
        \EndIf
    \EndWhile


\State \textbf{End Procedure}

\end{algorithmic}
\end{algorithm}

\vspace{-0.75em}
Algorithm~\ref{alg:algorithm} describes the flow of information and control in our framework. The framework takes as input the state graph $\mathbf{G_s}$ and attribute graph $\mathbf{G_k}$, along with an input prompt \textit{(ip\_prompt)} that contains information about the class of tasks, an in-context example, and a query specifying the task the agent must perform. The LLM generates an action sequence $T$ (Line 3), which is refined to $T_{refined}$ using the knowledge in the KG (Line 4). If there are no unresolved mismatches between KG and LLM output ($\varepsilon_{unkn}$), the action sequence is executed, with the outcomes and errors collected for further analysis (Lines 5-7). Any unresolved mismatches or errors in outcome result in a feedback prompt to the LLM, leading to a new predicted sequence of actions $T^{'}$ (Lines 9-13, 19-23). If these mismatches and/or errors persist (beyond threshold $F_{max}$), the agent queries a human, which potentially leads to knowledge refinement, updating $\mathbf{G_k}$ and $\mathbf{G_s}$. After the expansion, the knowledge base is updated, and the refined action sequence is executed and evaluated (Lines 14-18, 24-26). This entire process is repeated until the tasks is completed or some threshold (e.g., time limit) is exceeded.

\section{Experimental Sections and Results}
\label{sec:exp-setup-results}
This section describes the experimental setup and the results of experimentally evaluating three hypotheses:
\begin{itemize}
\vspace{-0.75em}
\item[\textbf{H1:}] Combining generic prediction of action sequences from LLMs with KG-based specific prior knowledge improves performance compared with just LLMs.
\item[\textbf{H2:}] Soliciting and using human feedback as needed supports incremental knowledge revision and results in improved performance compared with not using human feedback.
\item[\textbf{H3:}] Our framework adapts to new classes of tasks through incremental and transparent knowledge refinement.
\vspace{-0.5em}
\end{itemize}

\begin{figure}[tb]
\centering
\captionsetup{font=scriptsize}
\setlength{\belowcaptionskip}{-10pt}
\includegraphics[width=0.45\textwidth]{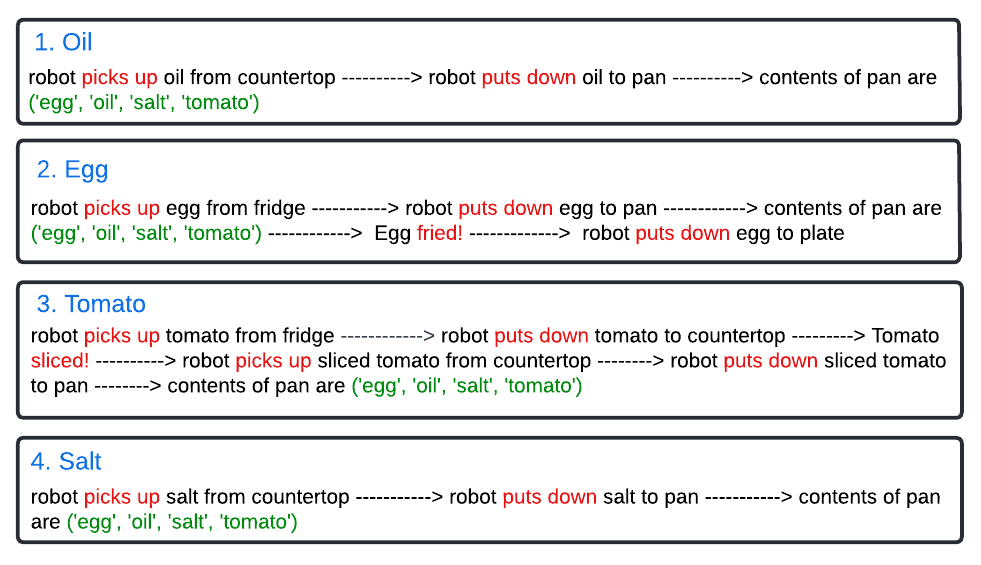}
\vspace{-0.75em}
\caption{Progress line~\cite{sakib2024cooking} showing use of each ingredient when preparing an \textit{omelette}.}
\vspace{-6pt}
\label{fig:progress-lines}
\end{figure}

\subsection{Experimental Setup}
\label{sec:expres-setup}
\noindent
We begin by describing the experimental set up, which includes the prompting of LLMs, and the choice of baselines, classes of tasks, and evaluation measures.

\noindent
\subsubsection{LLM Prompting}

We used GPT-3.5 and GPT-4o to generate action sequences for specific tasks in a given environment. We used Chain of Thought (CoT) prompting to encourage the LLM to decompose any given task into a series of logical steps. The main prompt included domain-specific information (e.g., object classes from $\mathbf{G_{s}}$), set $\mathbf{A}$ of agent's actions, and output for a single in-context example. For example, for the agent assisting in cooking tasks, the LLM was encouraged to generate an action sequence that fetches ingredients and tools, completes the cooking process, and serves the dish, based on the example of preparing coffee. 
The LLM's output was filtered to retain only the predicted action sequence.
As described in Section~\ref{sec:framework-hitl}, any unresolved mismatch between LLM and KG, or an error in outcome, also led to the agent sending a feedback prompt to the LLM in an attempt to fix the error.


\subsubsection{Baselines}
We evaluated three different configurations of components in our framework: (a) LLM; (b) LLM with a KG; and (c) LLM with KG and human input (LLM + KG + Human). We conducted linked trials, i.e., in each trial, the same LLM output was provided to each configuration. 
As stated in Section~\ref{sec:framework}, with just LLM, the predicted action sequence is sent directly for execution, and errors results in a feedback prompt to the LLM for a fixed number of times. 
With \textit{LLM + KG}, the KG is used to identify and fix mismatches between LLM output and KG; however, consistent mismatches and incorrect execution outcomes are not addressed. The \textit{LLM+ KG + human} configuration represents our framework, in which unresolved mismatches are addressed using human input, which is assumed to be accurate; the other two configurations serve as baselines.

\subsubsection{Classes of tasks}
In order to evaluate the ability of our framework to adapt to different classes of tasks, we considered cooking and cleaning tasks. Specifically, we considered $30$ different cooking tasks in a kitchen; this is the motivating scenario described in Section~\ref{sec:framework}. These tasks were created by sampling from the Recipe1M+ dataset~\cite{marin2021recipe1m+}. In addition, we considered $12$ variants of cleaning/clearing tasks that involved the agent cleaning specific objects or surfaces (e.g., "do the laundry"), or arranging objects in desired configurations in particular rooms (e.g., "clear the toys from the playroom")---see Figure~\ref{fig:12} for some examples.
The results of evaluating the adaptability of our framework is summarized later in Table~\ref{tab:h3_other_domain}.

\begin{figure}[tb]
\captionsetup{font=scriptsize}
\centering
    \begin{minipage}{0.45\textwidth}
        \begin{lstlisting}[basicstyle=\ttfamily\scriptsize]
'clean the bedroom_floor',
'dust the TV',
'wash the clothes',
'wash the dishes',
'water the plants',
'take out trash',
'clean the window',
'mop the countertop',
'clean the table',
'pick up and put all the toys in the toy box',
'charge the phone',
'play the music'
        \end{lstlisting}
    \end{minipage}
    \setlength{\abovecaptionskip}{-2pt}
    \setlength{\belowcaptionskip}{-12pt}
    \caption{12 variants of tasks that involve the agent assisting with cleaning different objects and surfaces, or clearing objects to achieve the desired object configuration.}
    \label{fig:12}
\end{figure}

\subsubsection{Evaluation Strategy}
For the evaluation of our framework, we used human participants to provide ground truth. Specifically, we recruited 18 human evaluators to mark the execution outputs for each task assigned to the framework and the two baselines. These evaluators were subject matter experts who were not involved in the design of our framework. The tasks were distributed such that the output for each task was evaluated by at least three human participants. The scores provided by the human (on a linear scale between 0-20) were averaged to obtain the success rate of our framework and the two baselines. These results are discussed further in Section~\ref{sec:expres-results}.

\vspace{-0.75em}
To better understand the LLM's performance, we also considered \textit{progress lines}~\cite{sakib2024cooking}, which depicted the use of key individual objects during individual steps of the action sequence, e.g., Figure~\ref{fig:progress-lines} shows the movement of each ingredient when cooking an omelette. These were presented along with the execution outputs to be evaluated by the humans.

\begin{table*}[tb]
\centering
\captionsetup{font=scriptsize}
\setlength{\belowcaptionskip}{-7pt}
\begin{tabular}{| >{\centering\arraybackslash} m{2.5cm}| 
>{\centering\arraybackslash} m{4cm}| >{\centering\arraybackslash} m{2.0cm}| >{\centering\arraybackslash} m{2.0cm}| >{\centering\arraybackslash} m{2.5cm}| }
\hline
LLM Models $\downarrow$ & Frameworks $\rightarrow$ & LLM & LLM + KG & LLM + KG + Human\\
\hline
\\[-1em]
 \multirow{4}{1.5cm}{GPT 4o}& \textbf{Success Rate (in \%) $\uparrow$} & 45.2 & 56.95 & \textbf{91.14} \\
\cline{2-5}
\\[-1em]
 & Avg. Tokens Used $\downarrow$ & 8316 & 7591 & 6459 \\
\cline{2-5}
\\[-1em]
&  Mean Ingd. Overlap (in \%) & 56.7 & 65.27 & \textbf{92.07} \\
\cline{2-5}
\\[-1em]
&  (\#nodes, \#edges) in $\mathbf{G_{s}}$ and $\mathbf{G_{k}}$ & (79, 772) & (79, 772) & \textbf{(87, 845)} \\
\hline
\\[-1em]
 \multirow{4}{1.5cm}{GPT 3.5}&  \textbf{Success Rate (in \%) $\uparrow$} & 25.41 & 33.95 & \textbf{92.08} \\
\cline{2-5}
\\[-1em]
 & Avg. Tokens Used $\downarrow$ & 8402 & 8415 & 4354 \\
\cline{2-5}
\\[-1em]
&  Mean Ingd. Overlap (in \%) $\uparrow$ & 38.13 & 44.29 & \textbf{98.97} \\
\cline{2-5}
\\[-1em]
&  (\#nodes, \#edges) in $\mathbf{G_{s}}$ and $\mathbf{G_{k}}$ & (79, 772) & (79, 772) & \textbf{(89, 869)} \\
\hline
\end{tabular}
\caption{Evaluating \textbf{H1 \& H2} for 30 recipes of six categories from Recipe1M+ dataset. The combination of LLM and KG ("LLM+KG") results in an increase in success rate, reduction in token use, and an increase in the mean ingredient overlap compared with just using the LLM ("LLM"). Also, soliciting and using human input when needed ("LLM+KG+Human") results in a substantial improvement on all measures, including an increase in the number of nodes and edges due to expansion of knowledge in KG.}
\label{tab:h2_kg_expansion}
\end{table*}

\begin{table*}[tb]
\centering
\captionsetup{font=scriptsize}
\setlength{\belowcaptionskip}{-7pt}
\begin{tabular}{| >{\centering\arraybackslash} m{2.5cm}| 
>{\centering\arraybackslash} m{4cm}| >{\centering\arraybackslash} m{2.0cm}| >{\centering\arraybackslash} m{2.0cm}| >{\centering\arraybackslash} m{2.5cm}| }
\hline
LLM Models $\downarrow$ & Frameworks $\rightarrow$ & LLM & LLM + KG & LLM + KG + Human\\
\hline
\\[-1em]
 \multirow{3}{1.5cm}{GPT 4o}&  \textbf{Success Rate (in \%) $\uparrow$} & 42.33 & 44.00 & \textbf{75.66} \\
\cline{2-5}
\\[-1em]
 & Avg. Tokens Used & 3820 & 3571 & 1979 \\
\cline{2-5}
\\[-1em]
&  (\#nodes, \#edges) in $\mathbf{G_{s}}$ and $\mathbf{G_{k}}$ & (39, 313) & (39, 313) & \textbf{(40, 331)} \\
\hline
\\[-1em]
 \multirow{3}{1.5cm}{GPT 3.5}&  \textbf{Success Rate (in \%) $\uparrow$} & 32.63 & 42.5 & \textbf{98.75} \\
\cline{2-5}
\\[-1em]
 & Avg. Tokens Used $\downarrow$ & 4963 & 4440 & 3510 \\
\cline{2-5}
\\[-1em]
&  (\#nodes, \#edges) in $\mathbf{G_{s}}$ and $\mathbf{G_{k}}$ & (39, 313) & (39, 313) & \textbf{(44, 397)} \\
\hline
\end{tabular}
\caption{Evaluating \textbf{H3} by adapting our framework to the cleaning and clearing tasks without requiring extensive tuning (e.g., of LLM) or comprehensive encoding of knowledge (in KG). We observed a substantial improvement on all performance measures with our framework compared with just using LLM outputs or even LLM+KG. Also, the agent is able to solicit human input as needed to incrementally and transparently revise knowledge in the KG.} 
\label{tab:h3_other_domain}
\end{table*}

\subsubsection{Evaluation Measures}
The key performance measures considered in this work include:
\begin{itemize}
    \vspace{-1em}
    \item \textbf{Success rate:} As stated above, this measure was computed based on the scores assigned by the human participants. Higher values are better as they indicate a higher degree of satisfaction in task completion.  This measure was used for evaluating H1-H3. 

    \item \textbf{Average tokens used:} The number of tokens used when prompting the LLM (including the input prompt and all subsequent feedback prompts) was averaged across all tasks. This is a measure of resource consumption and lower values are usually better, except when the use of prompts improves the values of other measures. 
    
    \item \textbf{Number of nodes and edges in KG:} We use this measure to evaluate H2 and H3. An increase in its value implies an expansion of knowledge in the KG.
    
    
    \item \textbf{Mean ingredient overlap:} A measure specific to the first class of tasks (cooking); it is the average overlap between the ingredients in the ground truth recipe and the ingredients in the executed action sequence. If $m_i$ denotes the ingredients required to make a particular dish and $l_i$ denotes the ingredients in the action sequence, this measure is computed as: 
    \vspace{-0.75em}   
    \begin{align}
        \label{eqn:mean-overlap}    
        \text{Mean ingredient overlap} = \frac{1}{N} \sum_{i=1}^{N} \frac{|m_i \cap l_i|}{|m_i|}
    \end{align}
    where $| \cdot |$ is the cardinality of a set, and $N$ is the total number of recipes sampled from the dataset.  This measure was used to evaluate H1-H2.   
    
\end{itemize}

\subsection{Experimental Results}
\label{sec:expres-results}
\noindent
Next, we describe and discuss the experimental results.

\vspace{-0.5em}
\noindent
\textbf{Evaluating H1.} 
We first explored whether the combination of LLM and KG leads to improved performance in comparison with just using LLM output for any given task. The corresponding results for the cooking-related tasks are summarized in Table~\ref{tab:h2_kg_expansion}; in particular, see columns labeled "LLM" and "LLM + KG". For the two LLMs considered (GPT3.5, GPT4o), we observe a substantial increase in success rate, reduction in token use, and an increase in the mean ingredient overlap for LLM+KG compared with LLM. These results provide strong support for \textbf{H1}.


\vspace{-0.5em}
\noindent
\textbf{Evaluating H2.}
Next, we explored the impact of soliciting and using human input as needed. The last column of Table~\ref{tab:h2_kg_expansion} ("LLM+KG+Human") shows that our framework's judicious use of human input with LLM and KG markedly improved performance on all measures compared with LLM and LLM+KG. With GPT-4o, we observed a $45.94\%$ increase in success rate over LLM and $34.19\%$ over LLM+KG. For GPT-3.5, the success rate increased by $66.67\%$ over LLM and $58.13\%$ over LLM+KG. Also, the average number of tokens used by our framework dropped by $48.26\%$ compared with baseline(s). This performance improvement was strongly influenced by the refinement of knowledge in the KG; the number of nodes and edges in the KG expanded from (79, 772) to (87, 845) with GPT-4o and to (89, 869) with GPT3.5. These results strongly support \textbf{H2}.

\vspace{-0.5em}
\noindent
\textbf{Evaluating H3.}
Finally, we evaluated the ability to adapt our framework to a different class of tasks (cleaning and clearing), with the results summarized in Table~\ref{tab:h3_other_domain}. Unlike prior work~\cite{sakib2024cooking}, we seek to achieve this adaptation without extensive tuning (e.g., of LLM) or the need for comprehensive domain-specific knowledge (in the KG). We instead leverage the interplay between LLM, KG, and human input to support incremental adaptation to the new class of tasks. Results indicate (once again) a substantial improvement on all measures for our framework compared with the baselines. We noted that the impact of adding different bits of knowledge to the KG can differ. For example, with GPT-4o, the addition of just one item (mopping\_cloth) to the KG based on human input led to a $31\%$ increase in success rate; with GPT3.5, this improvement was more pronounced ($56\%$). We also observed a substantial reduction in the number of tokens used. In addition, this adaptation of knowledge is fully transparent by design. These results strongly support hypothesis \textbf{H3}.

\section{Conclusions and Future Work}
\label{sec:conclusions}
Embodied agents assisting humans frequently have to complete previously unseen tasks or operate in new scenario. This paper describes a framework that leverages the complementary strengths of Large Language Models (LLMs), Knowledge Graphs (KGs), and Human-in-the-Loop (HITL) feedback to satisfy this requirement. Specifically, the generic task decomposition ability of LLMs is used to predict a sequence of abstract actions to complete any given task. This sequence is adapted to the specific scenario(s) and the task-, agent-, or domain-specific constraints using a KG that encodes prior knowledge of some objects, object attributes, and action capabilities. Any unresolved mismatch between the KG and the LLM output, and any unexpected action outcomes, are addressed by soliciting and using human input. This HITL feedback corrects errors and refines the existing knowledge (in the KG) for subsequent operation. Experimental evaluation in two simulated domains demonstrates substantial performance improvement compared with baselines, and illustrates incremental acquisition of knowledge to adapt to new classes of tasks.


\vspace{-0.75em}
This research opens up multiple avenues for further research. First, we will explore the use of this framework in many more classes of tasks, building on (and reinforcing) the promising results obtained so far.  Second, we will investigate the trade-off between automating the generation of an action sequence for any given task, and soliciting and incorporating human feedback as needed. Furthermore, we will explore the use of this framework on a physical robot platform assisting humans. The long-term objective is to create assistive agents and robots that can interact and collaborate with humans in different application domains.
\balance

\bibliographystyle{IEEEtran}

\end{document}